\documentclass[letterpaper]{article} 
\usepackage{aaai2026}  
\usepackage{times}  
\usepackage{helvet}  
\usepackage{courier}  
\usepackage[hyphens]{url}  
\usepackage{graphicx} 
\usepackage{natbib}  
\usepackage{caption} 
\newcommand{\myparagraph}[1]{\smallskip \noindent{\bf {#1}.}}
\usepackage{multirow}
\usepackage{booktabs} 
\usepackage{amsmath} 
\usepackage{algorithm}
\usepackage{algorithmic}
\usepackage{amssymb}

\usepackage{newfloat}
\usepackage{listings}
\DeclareCaptionStyle{ruled}{labelfont=normalfont,labelsep=colon,strut=off} 
\floatstyle{ruled}
\newfloat{listing}{tb}{lst}{}
\floatname{listing}{Listing}
\title{FastAnimate: Towards Learnable Template Construction and Pose Deformation for Fast 3D Human Avatar Animation}
\vspace{-2cm}
\author {
    Jian Shu$^*$,
    Nanjie Yao$^*$,
    Gangjian Zhang,
    Junlong Ren,
    Yu Feng,
    Hao Wang$^\dagger$
}
\affiliations {
    The Hong Kong University of Science and Technology (Guangzhou)\\
    jshu704@connect.hkust-gz.edu.cn, 
    nanjieyao@gmail.com, 
    gzhang292@connect.hkust-gz.edu.cn,  
    jren686@connect.hkust-gz.edu.cn,
    yufeng9819@gmail.com,
    haowang@hkust-gz.edu.cn
}

\begin{document}

\maketitle
\renewcommand{\thefootnote}{}
\footnotetext{\noindent $\dagger$: Corresponding author} 
\footnotetext{\noindent *: Equal contribution}

\begin{abstract}
3D human avatar animation aims at transforming a human avatar from an arbitrary initial pose to a specified target pose using deformation algorithms. 
Existing approaches typically divide this task into two stages: canonical template construction and target pose deformation. However, current template construction methods demand extensive skeletal rigging and often produce artifacts for specific poses. Moreover, target pose deformation suffers from structural distortions caused by Linear Blend Skinning (LBS), which significantly undermines animation realism. To address these problems,  we propose a unified learning-based framework to address both challenges in two phases. For the former phase, to overcome the inefficiencies and artifacts during template construction, we leverage a U-Net architecture that decouples texture and pose information in a feed-forward process, enabling fast generation of a human template. For the latter phase, we propose a data-driven refinement technique that enhances structural integrity.
Extensive experiments show that our model delivers consistent performance across diverse poses with an optimal balance between efficiency and quality, surpassing state-of-the-art (SOTA) methods.
\end{abstract}

\section{Introduction}


3D human avatar animation aims to deform an individual’s 3D model into a specified pose. The main challenge in 3D human animation is balancing high realism with real-time performance, personalization and natural motion. 
Prevailing approaches generally split the animation process into two phases: canonical template construction and target pose deformation~\cite{mcmanus2011influence, ichim2015dynamic, li20193d}. 
The first stage focuses on constructing high-quality, customizable human templates based on the provided body data; while the second stage seeks to minimize structural errors resulting from pose changes and reduce texture artifacts.

Prior studies on canonical template construction can be categorized into two approaches: pretrained general template and data-driven personalized template construction.
Pretrained models such as SCAPE~\cite{scape2005} and Skinned Multi-Person Linear (SMPL) model~\cite{loper2023smpl} use predefined mesh structures and statistical models from 3D scan data to represent human shape and pose. 
In contrast to traditional methods, 3D Gaussian-based methods~\cite{liu2024animatable, kocabas2024hugs, moreau2024human, pang2024ash, yuan2024gavatar} have emerged as a promising paradigm, representing human bodies as collections of anisotropic 3D Gaussians optimized for both rendering and animation. However, these require time-consuming data preprocessing for  which is not generalizable and struggle to manage self-intersecting regions on human bodies~\cite{kwon2024generalizable,moon2024expressive,wen2025life}. 



\begin{figure}[t]
\centering
    \includegraphics[width=0.95\columnwidth]{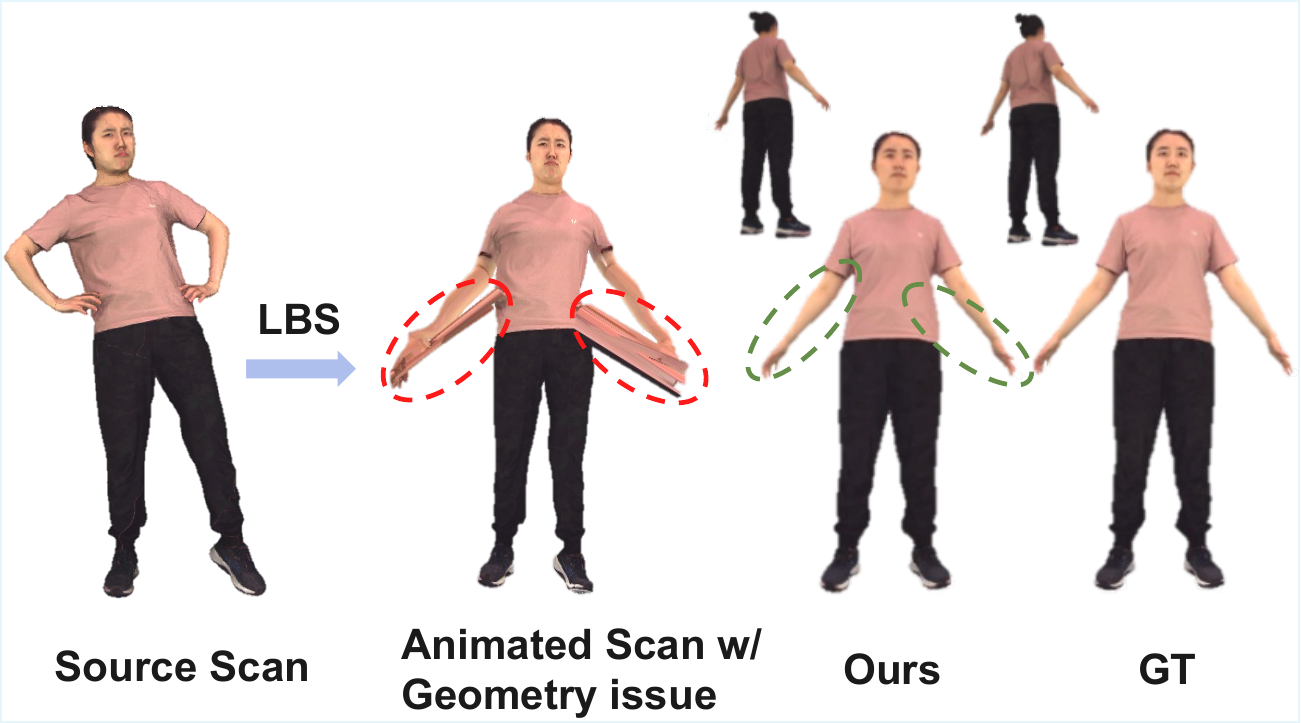}
    \caption{Visualization of the stretch-geometric problem under deformation (\textit{Left}) and our result (\textit{Right}).}
    \label{fig:intro}
    \vspace{-0.5cm}
\end{figure}

In terms of the target pose deformation phase, 
it greatly determines the quality of pose transformation. 
Linear Blend Skinning (LBS) allows vertices to be influenced by multiple skeletons through weighted averaging, offering a balance between efficiency and flexibility. However, it suffers from volume loss and limited capacity for non-linear deformations~\cite{james2005skinning, yang2006curve}. To address these limitations, Dual Quaternion Skinning~\cite{kavan2007skinning} replaced linear matrix blending with dual quaternions, helping to preserve volume and improve joint realism. Despite this, a key issue in this phase is that significant deformation discrepancies in joint regions can lead to geometric problems, for example, stretching or collapsing of the human mesh, illustrated in Figure~\ref{fig:intro} (\textit{Left}). This occurs because the commonly used LBS algorithm calculates vertex positions through a linear blend of joint transformations, without any inherent constraints on human body topology~\cite{kavan2008geometric, kavan2010fast, nuvoli2022skinmixer}. 



In this paper, we propose a learning-based Gaussian animation method, which contains the decoupled template representation and the learnable Gaussian animation modules. In the canonical template construction stage, we extract UV and pose features from human scan and SMPL-X body mesh respectively. Then these features are fed into a U-Net to initialize the parameters of human Gaussians. As a feed-forward process that directly learned from the raw source data, it is free from data preprocessing and only cost less than 50 ms for inference, which is much more efficient than the existing methods~\cite{ho2023custom, shen2023xavatar, moon2024expressive, wen2025life}. 


In the target pose deformation stage, we introduce a pretrained model as a strong human prior for realistic human animation. This model eliminates unexpected geometry issues (such as incorrect stretching) using geometry supervision losses and, based on the personalized template constructed earlier, corrects texture issues (like edge texture fusion) via color loss, with example results shown in Figure~\ref{fig:intro} (\textit{Right}).
Extensive experiments show that our proposed two-stage framework effectively prevents quality degradation from multi-stage error accumulation, outperforming current state-of-the-art (SOTA) approaches and demonstrating impressive efficiency advantage. The key contributions can be summarized as follows:
\begin{itemize}
    \item We propose an unified human animation framework, FastAnimate to realize pose-conditioned textured human avatar animation with balanced efficiency and fidelity. The entire inference process costs only about 0.1s.

    \item We propose a decoupled method to construct canonical human Gaussian templates. This approach separates UV feature and pose feature, eliminating the need for heuristic texture estimation.
    
    \item We design a data-driven, learnable Gaussian animation module capable of addressing geometry challenges arising from human deformation. This approach mitigates binding inaccuracies while preserving fine-grained geometry details.

\end{itemize}
\section{Related Work}
\begin{figure*}[ht]
\centering
    \includegraphics[width=0.975\textwidth]{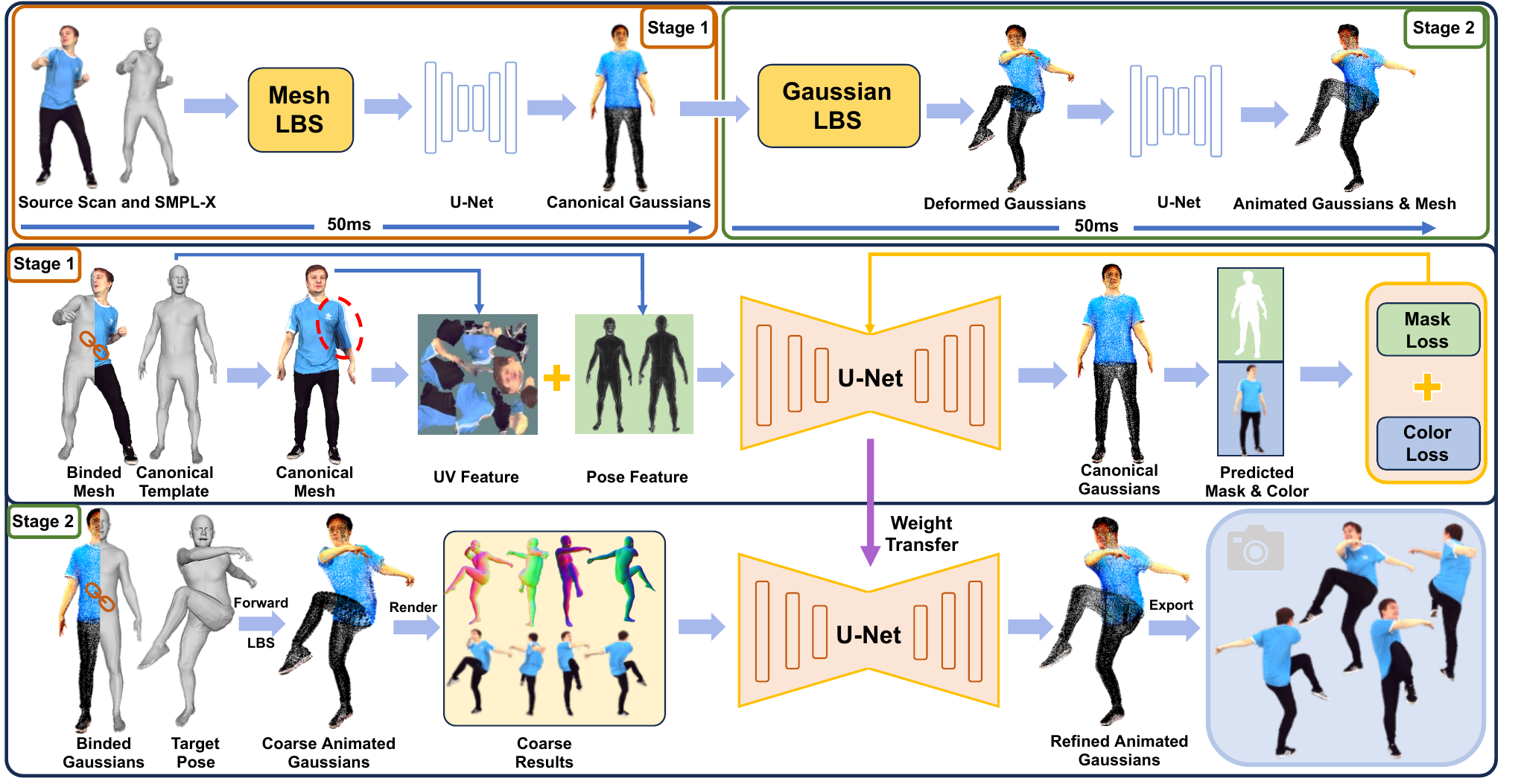}
    \vspace{-0.1cm}
    \caption{Overview of the proposed FastAnimate. 
    This framework consists of two stages: 
    In the first stage, we decouple the UV feature and pose feature from canonical mesh to build canonical Gaussians with the given human scans and SMPL-X canonical template. In the second stage, we utilize the LBS to drive canonical Gaussians to form coarse animated Gaussians. To further improve the animation quality, we utilize a coarse geometry refinement module to obtain high-fidelity refined animated human Gaussians. By leveraging FastAnimate, we can achieve robust 3D human animation results with enhanced texture quality and pose correctness.
    }
    \vspace{-0.4cm}
    \label{fig:pipeline}
\end{figure*}
\myparagraph{3D Avatar Animation} 
Animatable templates and animation algorithms are essential for 3D Human Avatar Animation. Early skeleton-based methods rig precise skeletons and weight surface points ~\cite{lewis2023pose, pons2015dyna}. Parametric human models expanded animatable template research ~\cite{scape2005, hasler2009bodyshape}, focusing on templates easily deformed via joint transformations ~\cite{pantuwong2012novel, fechteler2016real, feng2015avatar}. While SMPL enables deformation, it lacks photorealism ~\cite{varol2018bodynet}. Subsequent studies use SMPL models as 3D priors, reconstructing clothed avatars from monocular/sparse inputs ~\cite{aliakbarian2023hmd, xu2024relightable, lu2024avatarpose, karthikeyan2024avatarone}, but require specific data formats. Recently, mesh-based animation avoids complex preprocessing ~\cite{Saito:CVPR:2021, ho2023custom}, deforming meshes while preserving details. Based on the high-quality template, human avatar could be easily animated using LBS algorithm.  Our proposed method is inspired by these mesh-based approaches and goes further in both of the two main areas of avatar animation.

\myparagraph{Gaussian Splatting for Humans} Compared to previous representations, 3D Gaussian Splatting (3DGS) introduced a novel paradigm for human representation and rendering, utilizes explicit Gaussian ellipsoids to achieve high-quality novel view synthesis with real-time performance~\cite{kerbl20233dgs, zheng2024gps, zhang2024multigo, 11209141, zhang2025sat}. 3DGS offers a differentiable and efficient pipeline, making it a promising tool for digital human modeling. In human reconstruction, 3DGS enables the creation of detailed avatars from multi-view images or sparse point clouds with remarkable efficiency. For instance,~\cite{abdal2024gaussian} demonstrated that 3DGS can reconstruct photorealistic human models in minutes. Specifically, ~\cite{liu2024humangaussian} introduces structure-aware score distillation sampling to optimize the appearance and geometry of human gaussians. These studies widely explore the potential of 3DGS on human avatars. For human avatar animation, 3DGS has been adapted to support animation by integrating parametric models like SMPL. GauHuman~\cite{hu2024gauhuman} leverages LBS to deform Gaussians from a canonical pose to animated states, achieving high training and rendering speeds. Furthermore, through~\cite{kocabas2024hugs} real-time interaction is made feasible, which drive digital humans from monocular video inputs, capturing intricate details such as clothing wrinkles and hair motion with minimal latency. Impressed by the outstanding performance of 3D Human Gaussians, our research utlizes this representation as the base of our human model.

\section{Methodology}

\subsection{Preliminaries}
\myparagraph{SMPL-X Model} The SMPL~\cite{loper2023smpl} model is a parametric model that represents 3D human body shapes and poses. Our method is build upon one of its variants, SMPL-X, which can be denote as $\mathcal{X}(\theta, \beta, \alpha, \sigma) \in \mathbb{R}^{10475\times3}$, where the $\theta \in \mathbb{R}^{63}$ controls the global orientation and relative rotations of the body joints, the $\beta \in \mathbb{R}^{10}$ represents the body shape, and the $\alpha \in \mathbb{R}^{20}$ and $\sigma \in \mathbb{R}^{20}$ control the facial expression and finger movements, respectively. 

\myparagraph{3D Gaussian Splatting} We utilize 3D Gaussians~\cite{kerbl20233dgs}, an explicit and expressive differentiable 3D representation to model the textured and animated humans. Each human is represented as a set 3D Gaussians $\mathcal{G} = \{\mathcal{G}_i\}$, where each 3D Gaussian is parameterized as: $\mathcal{G}_i\ = \{\mu_i, \alpha_i, s_i, r_i, \textbf{c}_i\} \in \mathbb{R}^{14}$. The $\mu_i \in \mathbb{R}^3$ is the geometric center, $\alpha_i \in \mathbb{R}^1$ is the opacity value, $s_i \in \mathbb{R}^3$ and $r_i \in \mathbb{R}^4$ denotes the scale and rotation parameters, respectively, and the $\textbf{c}_i \in \mathbb{R}^3$ is the RGB color. This parametric formulation compactly encodes both spatial properties and visual attributes while maintaining differentiability.

\myparagraph{Linear Blend Skinning} LBS is a widely adopted technique in character animation and 3D modeling for deforming a body with skeletal structure. Given its computational efficiency and compatibility with parametric models, LBS serves as a foundational component in our proposed framework. Here, we briefly outline its formulation and key properties to provide context for subsequent sections.

In this study, we consider a 3D mesh \(\mathcal{M} = \{\mathbf{v}_i\}_{i=1}^V\) with \(V\) vertices, where \(\mathbf{v}_i \in \mathbb{R}^3\) represents the position of the \(i\)-th vertex in a rest pose, alongside a 3D Gaussian representation. Both are rigged to a skeleton from the SMPL-X model, which includes \(J\) joints, each defined by a transformation matrix \(\mathbf{T}_j \in \mathbb{R}^{4 \times 4}\) capturing rotation and translation relative to the rest state. The LBS algorithm deforms the mesh by calculating transformed vertex positions \(\mathbf{v}_i'\) as:

\begin{equation}
\mathbf{v}_i' = \sum_{j=1}^J w_{i,j} \mathbf{T}_j \mathbf{v}_i,
\end{equation}
where \(w_{i,j} \in [0, 1]\) denotes the blend weight of the \(j\)-th joint on the \(i\)-th vertex, with \(\sum_{j=1}^J w_{i,j} = 1\) for normalization. Here, \(\mathbf{T}_j\) is derived from SMPL-X models fitted to two distinct poses, and the weights \(\mathbf{W} = \{w_{i,j}\}\) are pre-computed during rigging, based on vertex-joint proximity or data-driven optimization.
In the following sections, we build upon this mechanism to enhance the geometric and visual fidelity of animated human avatar representations.

 
\subsection{Method Overview}
To achieve robust and generalizable human avatar animation, we propose a learning-based framework with symmetric forward-backward architecture, an approach that leverages a two-stage transformation process with transferred weights. This method ensures pose alignment while preserving texture integrity and enables efficient animation to a target pose. The methodology is detailed as follows:

Given an input human avatar representation \(\mathcal{A}\), characterized by its 3D geometry \(\mathcal{M}\) for clothed human pose/shape and 2D texture \(\mathcal{T}\) for clothed human appearance, we first transform it into a pre-defined intermediate canonical pose using the forward component of the symmetric structure. This step aims to standardize the pose information while retaining accurate texture details. Formally, let \(\mathcal{P}_u\) denote the canonical pose, and \(\mathcal{F}_{\text{fwd}}(\cdot)\) represents the forward transformation function parameterized by a learned module with weights \(\mathbf{W}\). The intermediate representation \(\mathcal{A}_u\) is computed as:
\begin{equation}
\mathcal{A}_u = (\mathcal{M}_u, \mathcal{T}_u) = \mathcal{F}_{\text{fwd}}(\mathcal{A}, \mathcal{P}_u; \mathbf{W}),
\end{equation}
where \(\mathcal{M}_u\) is the geometry aligned to \(\mathcal{P}_u\), and \(\mathcal{T}_u\) approximates the original texture \(\mathcal{T}\) with minimal distortion. The forward transformation leverages SMPL-X geometric priors and texture mapping to ensure consistency.

Subsequently, the backward component of the symmetric structure transforms \(\mathcal{A}_u\) into the target pose \(\mathcal{P}_t\). This process is regarded as the inverse of the forward transformation, defined by the function \(\mathcal{F}_{\text{bwd}}(\cdot)\), which uses the fine-tuned \(\mathbf{\widetilde{W}}\) shared from \(\mathbf{W}\) due to the symmetry of the framework. The final animated avatar \(\mathcal{A}_t\) is obtained as:
\begin{equation}
\mathcal{A}_t = (\mathcal{M}_t, \mathcal{T}_t) = \mathcal{F}_{\text{bwd}}(\mathcal{A}_u, \mathcal{P}_t; \mathbf{\widetilde{W}}),
\end{equation}
where \(\mathcal{M}_t\) corresponds to the target geometry, and \(\mathcal{T}_t\) preserves the texture aligned with \(\mathcal{P}_t\). The symmetry implies that \(\mathcal{F}_{\text{bwd}}\) mirrors \(\mathcal{F}_{\text{fwd}}\) in structure but operates in the opposite direction, \(\mathcal{F}_{\text{bwd}} \approx \mathcal{F}_{\text{fwd}}^{-1}\), constrained by the shared parameterization.

The transfer of weights between the two models offers two key advantages. Firstly, by sharing weights \(\mathbf{W}\) between \(\mathcal{F}_{\text{fwd}}\) and \(\mathcal{F}_{\text{bwd}}\), the framework reduces the number of trainable parameters and enforces consistency across the transformation pipeline. Secondly, this structure enhances the generation ability of the model under limited access to related dataset. During training, the loss function comprises two components: one for geometry and one for texture. The geometry loss computes the mean squared error (MSE) between the predicted mask \(\mathcal{K}^p\) and ground truth mask \(\mathcal{K}^g\), summed over n viewpoints, while the texture loss calculates the MSE between the predicted color \(\mathcal{C}^p\) and ground truth color \(\mathcal{C}^g\), also summed over n viewpoints. The total loss is:
\begin{equation}
\mathcal{L} = \sum_{v=1}^{n} \|\mathcal{K}^{p,v} - \mathcal{K}^{g,v}\|_2^2 + \sum_{v=1}^{n} \|\mathcal{C}^{p,v} - \mathcal{C}^{g,v}\|_2^2,
\end{equation}
where \(v\) denotes the viewpoint index. This formulation ensures effective optimization of both geometric accuracy and texture fidelity across multiple perspectives.

\subsection{Decoupled Template Representation}
In this section, we introduce the method of generating Decoupled Template Representation. This method utilizes a human mesh in arbitrary pose at the beginning and transforms it to the canonical space with pose prior extracted from corresponding SMPL-X model. 
To enable efficient and pose-driven human animation, we propose a novel method termed \textit{Decoupled Template Representation}. This method decouples human texture and geometry, extracting information from UV features and pose features to achieve a standardized, animatable 3D explicit representation based on 3D Gaussians. This animatable template could be built with the following steps:

Given a source human scan mesh \(\mathcal{M}_s\), skeletal rigging is performed by binding it to the source SMPL-X template \(\mathcal{X}_s\), which is parameterized by shape \(\beta_s\) and pose \(\theta_s\). The binding process employs LBS with pre-defined blend weights \(\mathcal{W}\), associating each vertex of \(\mathcal{M}_s\) with the skeletal joints of \(\mathcal{T}_s\). Next, the LBS transformation matrix \(\mathbf{T}_{s \to c}\) that maps the source SMPL-X template \(\mathcal{X}_s\) to its canonical SMPL-X counterpart \(\mathcal{X}_c\) is computed, where \(\mathcal{T}_c\) is defined in a A-pose  with identical shape parameters. The transformation matrix from the source to the A-pose is derived as:
\begin{equation}
\mathbf{T}_{s \to c} = \sum_{i} \mathcal{W}_i \cdot \mathbf{G}_i(\theta_c, J(\beta_s)) \cdot \mathbf{G}_i^{-1}(\theta_s, J(\beta_s)),
\end{equation}
where \(J(\beta_s)\) denotes the joint locations regressed from the shape parameters, \(\mathbf{G}_i(\theta, J)\) represents the global transformation of the \(i\)-th joint, and \(\mathcal{W}_i\) are the skinning weights.

Applying \(\mathbf{T}_{s \to c}\) to the source mesh \(\mathcal{M}_s\), we obtain a coarse standardized human mesh \(\mathcal{M}_c\):
\begin{equation}
\mathcal{M}_c = \mathbf{T}_{s \to c} \cdot \mathcal{M}_s.
\end{equation}
This standardized mesh aligns with the canonical SMPL-X topology, facilitating subsequent processing. Then texture features \(\mathcal{F}_{\text{tex}}\) are extracted from UV maps using a pre-trained feature extractor, capturing appearance details such as color and surface patterns. These texture features are concatenated with pose-dependent features \(\mathcal{F}_{\text{pose}}\) derived from \(\mathcal{T}_c\), which encode geometric deformations induced by the canonical pose.

The combined feature set \(\mathcal{F} = [\mathcal{F}_{\text{tex}}, \mathcal{F}_{\text{pose}}]\) is fed into a U-Net architecture with reconstruction capabilities to initialize a set of 3D Gaussians \(\mathcal{G}\). Each Gaussian is parameterized by its mean \(\mu\), covariance \(\Sigma\), and opacity \(\alpha\), representing a local volumetric distribution aligned with the canonical SMPL-X template. The U-Net outputs the Gaussian parameters as:
\begin{equation}
\mathcal{G} = \{\mu_i, \Sigma_i, \alpha_i\}_{i=1}^N = \text{U-Net}(\mathcal{F}),
\end{equation}
where \(N\) is the number of Gaussians determined adaptively based on the mesh complexity.

Since the resulting Gaussians \(\mathcal{G}\) correspond to the canonical SMPL-X template \(\mathcal{X}_c\), animating the human representation becomes straightforward. Given a target SMPL-X model \(\mathcal{X}_t\) with a new pose \(\theta_t\), we compute the forward LBS transformation:
\begin{equation}
\mathbf{T}_{c \to t} = \text{LBS}(\mathcal{T}_t, J(\beta_s), \theta_t, \mathcal{W})\,  
\end{equation}
This transformation is applied to the Gaussian means:
\begin{equation}
\mu_i' = \mathbf{T}_{c \to t} \cdot \mu_i, \quad \forall i \in \{1, \dots, N\},
\end{equation}
while the covariance \(\Sigma_i\) and opacity \(\alpha_i\) remain unchanged, preserving the local structure and appearance. This process yields an decoupled animatable human template representation \(\mathcal{G}'\) driven efficiently by 3D Gaussians, suitable for real-time applications.

\subsection{Learnable Gaussian Animation}


In this section, we introduce the proposed approach to realize learning-based gaussian animation. 3D Human Gaussians are suitable for point-level deformation. However, incorrect geometry and unreal texture still exist as the animation matrix calculated by LBS may not be prefect when applied on mesh or Gaussians. 

To enhance the quality of coarse 3D Gaussian representations for human avatar animation, we introduce a method called \textit{Learnable Gaussian Animation}. The key component of this approach is a coarse geometry refinement module, which could refine an initial coarse 3D Gaussian model by leveraging geometric alignment with a target SMPL-X model, improving both fidelity and detail in the resulting human representation. The process is outlined as follows:

Starting with a coarse 3D Gaussian set \(\mathcal{G}_c = \{\mu_i, \Sigma_i, \alpha_i\}_{i=1}^N\), where \(\mu_i\), \(\Sigma_i\), and \(\alpha_i\) denote the mean, covariance, and opacity of each Gaussian, we render images from four distinct viewpoints \(\{V_1, V_2, V_3, V_4\}\). These rendered images, denoted as \(\mathcal{I}_c = \{I_c^{(k)}\}_{k=1}^4\), capture the coarse geometry from multiple perspectives. Concurrently, we render images \(\mathcal{I}_t = \{I_t^{(k)}\}_{k=1}^4\) from the same viewpoints using a target SMPL-X model \(\mathcal{T}_t\), parameterized by shape \(\beta_t\), pose \(\theta_t\), and expression \(\psi_t\). The SMPL-X model provides a high-fidelity pose reference.

For each viewpoint \(k\), we extract geometry features from both the coarse Gaussian renderings and the SMPL-X renderings. Let \(\mathcal{F}_c^{(k)} = \text{GeoEnc}(I_c^{(k)})\) and \(\mathcal{F}_t^{(k)} = \text{GeoEnc}(I_t^{(k)})\) represent the geometry features encoded by a shared geometry encoder \(\text{GeoEnc}(\cdot)\), which captures surface normals and silhouette information. These features are paired to form a combined feature set:
\begin{equation}
\mathcal{F}_{\text{pair}}^{(k)} = [\mathcal{F}_c^{(k)}, \mathcal{F}_t^{(k)}], \quad k = 1, 2, 3, 4.
\end{equation}
This paired feature set \(\mathcal{F}_{\text{pair}} = \{\mathcal{F}_{\text{pair}}^{(k)}\}_{k=1}^4\) integrates multi-view geometric cues from both the coarse and target representations.

The paired features \(\mathcal{F}_{\text{pair}}\) are input into a specialized U-Net architecture, designed to refine the coarse Gaussian parameters by leveraging the detailed geometric information from \(\mathcal{T}_t\). The U-Net outputs a refined Gaussian set \(\mathcal{G}_r\):
\begin{equation}
\mathcal{G}_r = \{\mu_i', \Sigma_i', \alpha_i'\}_{i=1}^{N'} = \text{U-Net}(\mathcal{F}_{\text{pair}}),
\end{equation}
where \(N'\) may differ from \(N\) due to pruning or addition of Gaussians. The refinement process involves two key operations: (1) pruning artifacts via a pre-defined Gaussian opacity threshold, such as redundant Gaussians causing visual noise, by adjusting \(\alpha_i'\) to suppress irrelevant regions; and (2) supplementing details, such as clothing wrinkles and finger geometry, by introducing new Gaussians or adjusting \(\mu_i'\) and \(\Sigma_i'\) to align with \(\mathcal{F}_t\). This alignment ensures that the refined Gaussians \(\mathcal{G}_r\) closely match the target SMPL-X geometry.

Additionally, the corrected geometric features from \(\mathcal{T}_t\) enable reinitialization of the Gaussians when necessary. The refined Gaussian parameters are computed as:
\begin{equation}
\mu_i' = \mu_i + \Delta\mu_i(\mathcal{F}_t), \quad \Sigma_i' = \Sigma_i + \Delta\Sigma_i(\mathcal{F}_t), \quad \alpha_i' = f(\alpha_i, \mathcal{F}_t),
\end{equation}
where \(\Delta\mu_i\), \(\Delta\Sigma_i\), and \(f(\cdot)\) are learned corrections derived from the U-Net which is pretrained as human reconstruction model, conditioned on the target SMPL-X features. This process yields a high-quality 3D Gaussian representation \(\mathcal{G}_r\), capable of supporting realistic human avatar animation with improved geometric fidelity and vivid texture transformation. We also observe that this module demonstrates excellent performance in correcting point-level geometric errors. As a result, this method could provide structurally accurate, high-fidelity human deformation results.
\section{Experiments}
\subsection{Experiment Setup}

\begin{figure*}[t!]
\centering
    \includegraphics[width=1\textwidth]{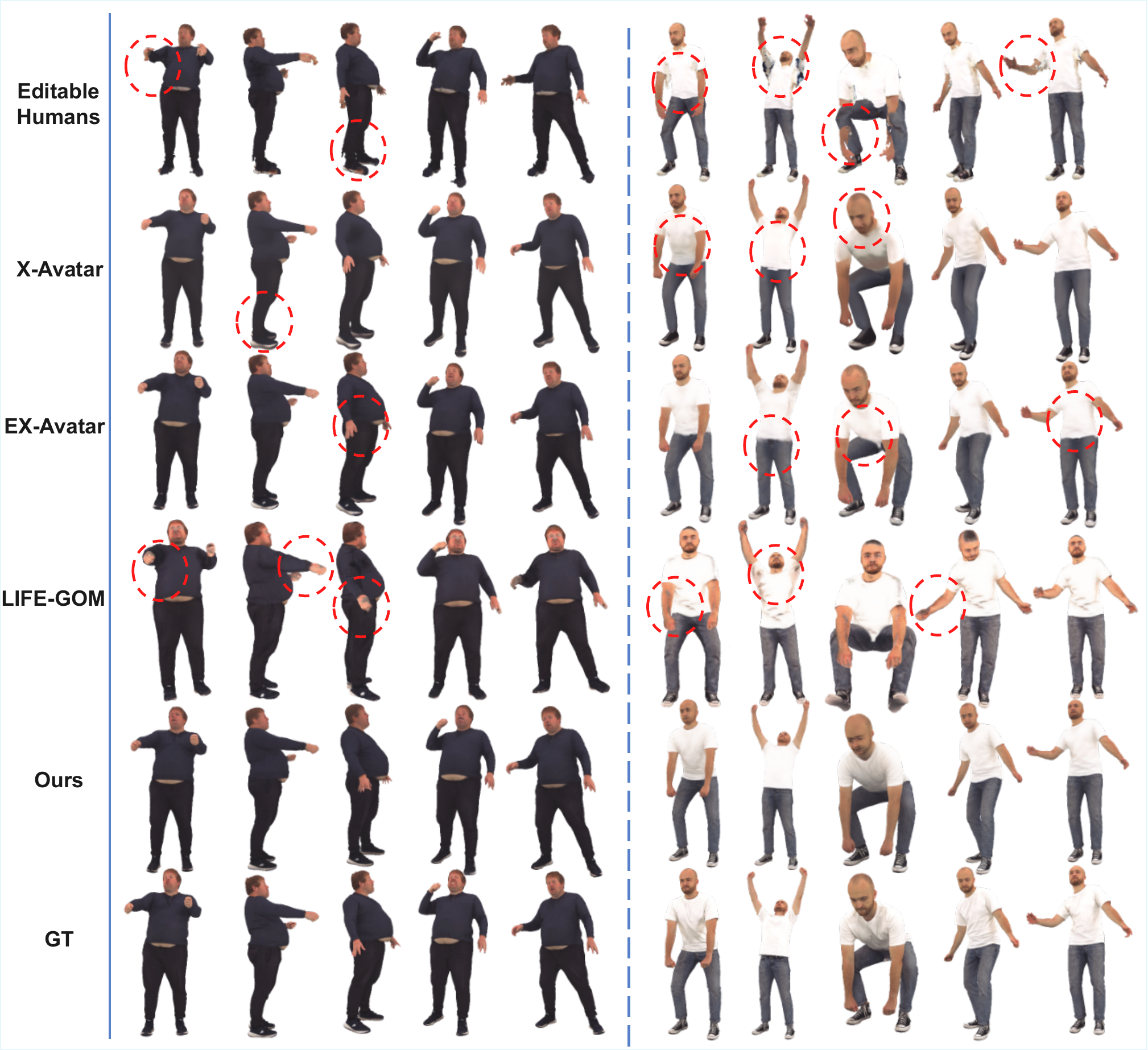}
    \vspace{-0.7cm}
    \caption{Qualitative comparison with state-of-the-art methods on novel pose synthesis. Please note that due to the difference in camera parameters, the results of LIFE-GOM has a marginal angle difference with others. Please zoom in for a detailed view.}
    \label{fig:experiment}
    \vspace{-0.25cm}
\end{figure*}


\myparagraph{Dataset} Our experiments is fully conducted on the X-Humans~\cite{shen2023xavatar} dataset. X-Humans dataset consists of 20 subjects with different garments. There are over 29K poses for training and 6.4K test poses. We follow the default setting to split the training and testing data. We sample 1000 scans from the training set as training data and follow EX-Humans~\cite{moon2024expressive} to select 20 scans from the test set.


\myparagraph{Baselines} To demonstrate the superiority of our proposed method, we compare the current SOTA 3D human animation baselines including: \textbf{Editable-Humans}~\cite{ho2023custom}, \textbf{X-avatar}~\cite{shen2023xavatar}, \textbf{EX-avatar}~\cite{moon2024expressive}, and \textbf{LIFE-GOM}~\cite{wen2025life}. These baselines represent distinct technical paradigms: Editable-Humans employs a 3D scan-based animation approach, while both X-Avatar and EX-Avatar utilize video-driven animation frameworks. The most similar approach to our method is LIFE-GOM, which establishes a 3D human Gaussian avatar from sparse-view inputs. See more details about baselines in Supplementary Material.

\myparagraph{Metrics} We follow the metrics used in Editable-humans~\cite{ho2023custom} to quantitatively evaluate the 3D pose correctness of animated human: Chamfer Distance (\textbf{CD}), Normal Consistency (\textbf{NC}), and \textbf{f-score}~\cite{tatarchenko2019single} and utilize the Peak Signal-to-Noise Ratio (\textbf{PSNR}), Structure Similarity Index Measure (\textbf{SSIM})~\cite{wang2004image} and Learned Perceptual Image Patch Similarity score (\textbf{LPIPS}) ~\cite{zhang2018unreasonable} to evaluate the 2D texture quality of animated human. Additionally, we utilize the inference \textbf{Time Cost} to assess the computational efficiency.

\subsection{Evaluation}

\myparagraph{Quantitative Evaluation} We present a comprehensive quantitative comparison of our method against SOTA animation approaches in Table~\ref{tab:cmp}. Our framework achieves superior performance across animated pose correctness, texture quality, and computational efficiency. For pose correctness, our method attains a CD of 0.609/0.701, NC of 0.934, and F-score of 86.118, outperforming existing techniques by significant margins. These metrics validate our approach’s ability to preserve topological fidelity during animation. In terms of texture quality, our method achieves a PSNR of 23.995/23.394~(F/B), SSIM of 0.9789/0.9838~(F/B), and LPIPS score of 0.0271/0.0358~(F/B) on rendered animated avatars. This demonstrates marked improvements in preserving high-frequency texture details and minimizing perceptual artifacts compared to prior work. In addition to computational efficiency, our framework animates 3D human scans in approximate 0.1s per instance—an order-of-magnitude speedup over existing methods without sacrificing output quality. These results collectively demonstrate that the superior performance of our framework on 3D human animation, simultaneously achieving unprecedented accuracy, visual quality, and practical deployability.


\begin{table*}[h]
\centering

\scalebox{0.8}{
\begin{tabular}{lc|ccc|ccc|c}
\toprule

\multirow{3}{*}{Methods} & \multirow{3}{*}{Publication} & \multicolumn{3}{c}{2D Texture Quality} & \multicolumn{3}{c}{3D Geometry Correctness} & Efficiency \\
                     ~     & ~  & \textbf{PSNR}: F/B $\uparrow$ & \textbf{SSIM}: F/B $\uparrow$ & \textbf{LPIPS}: F/B $\downarrow$ & \begin{tabular}{c} \textbf{CD}: P-to-S /\\ S-to-P $(\mathrm{cm}) \downarrow$ \end{tabular} & \textbf{NC$\uparrow$} & \textbf{F-score$\uparrow$} & \textbf{Time Cost$\downarrow$} \\
\midrule
Editable-Humans
& CVPR 2023 & 19.354/16.860 & 0.9302/0.9383 & 0.1019/0.1056 &  1.936/2.099  & 0.815  & 37.441 &   $\approx$22s    \\
X-Avatar
& CVPR 2023 & 20.824/19.335 & 0.9446/0.9443 & 0.0745/0.0791 &  0.975/0.924  & 0.907  & 65.986 & $\approx$10s       \\
EX-Avatar$^{\dagger}$
& ECCV 2024 & 22.201/21.984 & 0.9512/0.9443 & 0.0687/0.0712 &  0.732/0.717  & 0.921      & 84.231 & $\approx$10s \\
LIFE-GOM$^{\dagger}$
& ICLR 2025 & 22.103/22.294 & 0.9540/0.9521 & 0.0690/0.0687 &  0.801/0.772  & 0.923      & 83.204 & $\approx$1s \\
\midrule
FastAnimate $^{\dagger}$                               & AAAI 2026        & \textbf{23.995}/\textbf{23.394} & \textbf{0.9789}/\textbf{0.9838} & \textbf{0.0271}/\textbf{0.0358}  & \textbf{0.609}/\textbf{0.701} & \textbf{0.934} & \textbf{86.118} & \textbf{$\approx$0.1s} \\
\bottomrule
\end{tabular}
}
\caption{We compare our proposed method with SOTA approaches in terms of 2D Texture Quality (PSNR, SSIM, LPIPS), 3D Geometry Correctness (CD, NC, F-score) and Computational Efficiency (Time Cost). The ``$^{\dagger}$'' denotes the method is build upon the 3D Gaussian Splatting. For GS-based methods, the mesh are exported with technique provided by LGM~\cite{tang2024lgm} for fair comparison. The arrow $\uparrow$/$\downarrow$ represents the higher/lower is better.}

\label{tab:cmp}
\end{table*}

\begin{table}[!t]
\begin{center}
\scalebox{0.8}{
\begin{tabular}
{l|ccc}
\toprule

\multirow{2}{*}{Methods}  & \multicolumn{3}{c}{2D Texture Quality}  \\
&\textbf{PSNR}: F/B $\uparrow$ & \textbf{SSIM}: F/B $\uparrow$ & \textbf{LPIPS}: F/B $\downarrow$  \\
\midrule
w/o $LGA$      & 22.956/23.125 & 0.9721/0.9741 & 0.0472/0.0456
\\  
w/o $DTR$      & 23.183/23.284 & 0.9779/0.9830 & 0.0361/0.0366       \\ 
Full Pipeline & 23.995/23.394 & 0.9789/0.9838 & 0.0271/0.0358       \\  
\midrule

& \multicolumn{3}{c}{3D Geometry Correctness} \\
& \begin{tabular}{c} \textbf{CD}: P-to-S /\\ S-to-P $(\mathrm{cm}) \downarrow$ \end{tabular} & \textbf{NC} $\uparrow$ & \textbf{F-score}$\uparrow$   \\

\midrule
w/o $LGA$       & 0.823/0.857 & 0.902 & 82.524 \\  
w/o $DTR$       & 0.752/0.808 & 0.923 & 84.743 \\ 
Full Pipeline  & 0.609/0.701 & 0.934 & 86.118 \\    

\bottomrule
\end{tabular}
}
\caption{Ablation Study of Learnable Gaussian Animation module and Decoupled Template Representation. \label{tbl:abl}}
\vspace{-0.5cm}
\end{center}
\end{table}

\myparagraph{Qualitative Evaluation} Figure~\ref{fig:experiment} showcases our framework’s ability to generate animatable 3D human avatars with high-fidelity details across diverse poses. As demonstrated in Figure~\ref{fig:experiment}, our method generalizes robustly to novel poses while preserving intricate clothing folds, dynamic facial expressions, and articulated finger movements. Crucially, the framework remains stable even under extreme body posture changes (e.g., crouching, arm waving) where traditional methods typically fail, as evidenced by artifact-free deformations in challenging configurations. Figure~\ref{fig:experiment} provides a visual comparison with SOTA animation approaches. Editable Humans fails to reconstruct fine-grained hand geometry and introduces unnatural surface jaggedness. The results of X-Avatar and EX-Avatar exhibit visible degradation in clothing wrinkles and facial expressions due to methods' limitations. While LIFE-GOM addresses some topological issues, its reliance on input images leads to blurred textures in occluded regions, particularly around armpits, palms and backs of arms. These visualization results further demonstrate the superiority of our proposed FastAnimate. More results can be seen in the Supplementary Material.

\subsection{Ablation Study}
\begin{figure}[t]
\centering
    \includegraphics[width=0.95\columnwidth]{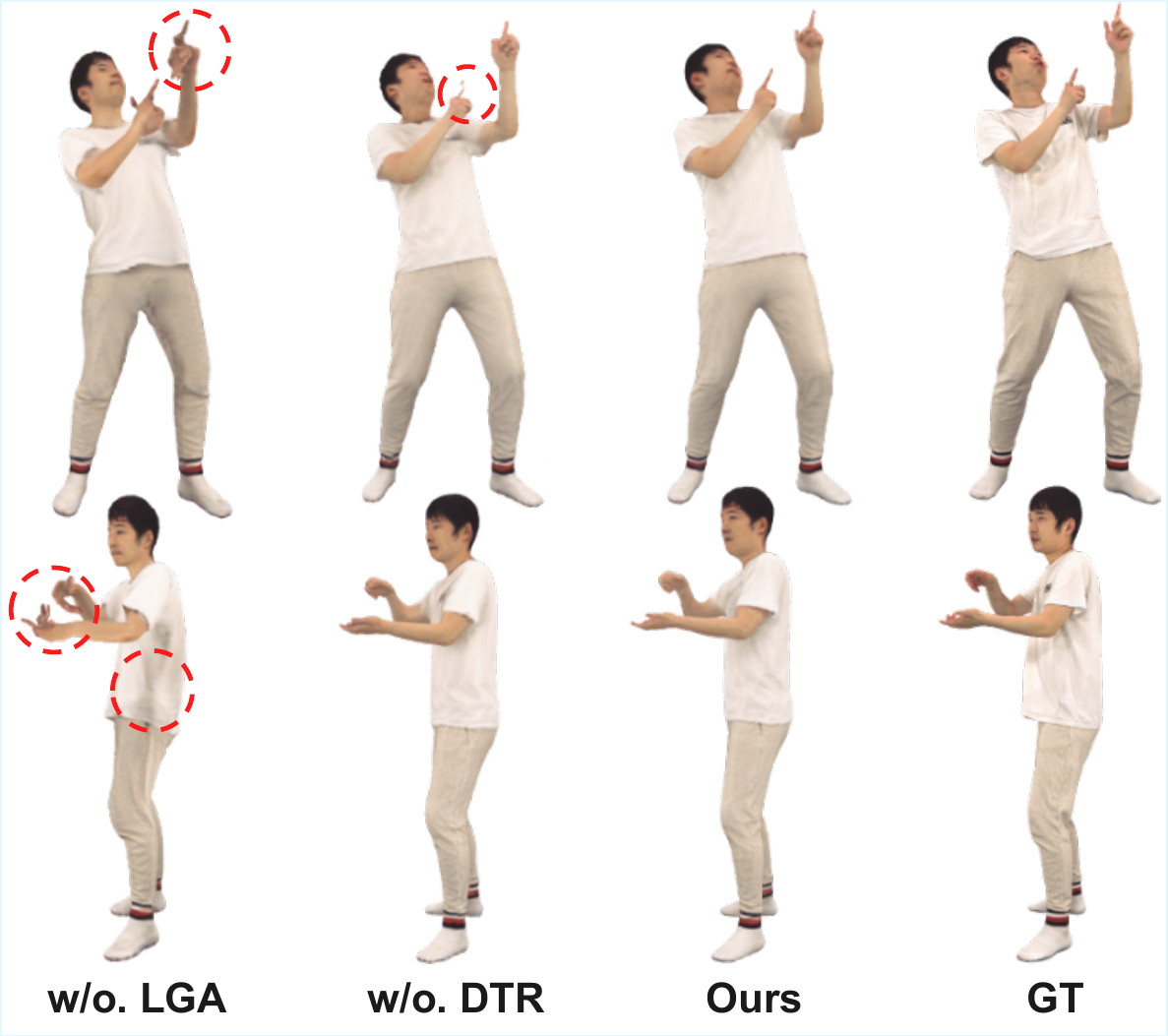}
    \caption{Visual ablation of proposed modules. Proposed modules improve fine-grained texture and geometry quality.}
    \label{fig:ablation}
    \vspace{-0.5cm}
\end{figure}

In this section, we conduct ablation studies to systematically analyze the impact of the components of our proposed method. We use a simple LBS approach as our baseline in ablation study. In the setting of ``w/o $LGA$'', we remove the entire coarse geometry refinement module in Learnable Gaussian Animation module to assess the impact. In the setting of ``w/o $DTR$'', we ablate the process of decoupled template representation and generate the animated 3D human Gaussian avatars with the U-Net directly.

\myparagraph{Effectiveness of Geometry Refinement} Table~\ref{tbl:abl} highlights the effectiveness of the proposed the learnable geometry refinement module. Ablating this component results in a significant decrease in both texture quality and geometry accuracy, underscoring its critical importance. We attribute this performance gap to two key factors: First, inaccurate binding of Gaussian to canonical template causes the misalignment in fine-grained geometry. Second, canonical poses fail to account for non-rigid deformations and high-frequency details inherent in dynamic human performances.  
The results validate that geometry refinement is indispensable for achieving photorealistic human animation.

\myparagraph{Effectiveness of Template Representation} Table~\ref{tbl:abl} illustrates
the effectiveness of our proposed template representation.  Removing this component causes a decrease in SSIM and PSNR metrics and increases LPIPS score, indicating severe quality degradation in texture quality. It is likely due to the high coupling between the UV feature and the pose feature, which forces the model to estimate texture details heuristically rather than leveraging disentangled, reusable templates. Furthermore, omitting the template framework introduces redundant computations: Without persistent template storage, the model must reprocess identical inputs across varying target poses. The ablation confirms that explicit template representation is crucial for both accuracy and computational efficiency in pose-transfer scenarios.

\myparagraph{Visual Ablation}
Figure~\ref{fig:ablation} demonstrates the perceptual and geometric improvements enabled by our proposed framework. The first column reveals significant degradation in animated avatar details when ablating the learnable geometry refinement module, manifesting as distortion in finger articulation and artifacts in waist clothing. The second column highlights minor yet perceptible losses in fine-grained appearance details, such as reduced fidelity in clothing wrinkles and residual finger inaccuracies. Strikingly, our full framework synthesizes avatars with high-fidelity textures and dynamically coherent wrinkles that align closely with ground truth observations, underscoring the necessity of both modules for photorealistic 3D human animation.
\section{Conclusion}
Real-time, high-quality human avatar animation under specified poses remains a persistent challenge in animation research, often yielding unrealistic results marred by significant artifacts and geometric errors. Traditional approaches prioritize improving animatable templates but overlook the animation process itself. In this study, we introduce a novel feed-forward framework that delivers easily animatable human Gaussians template while incorporating a dedicated module to refine problematic geometries. Extensive experiments demonstrate that the proposed FastAnimate generates canonical template and highly realistic reposed human avatars in approximately 0.1s, achieving an optimal balance between quality and speed.
\section{Acknowledgements}
This work is supported by the National Natural Science Foundation of China (No. 62406267), the Guangzhou-HKUST(GZ) Joint Funding Program (No. 2025A03J3956) and the Guangzhou Municipal Education Project (No. 2024312122).

\bibliography{LaTeX_CRC/main}

\end{document}